\documentclass[conference]{IEEEtran}
\IEEEoverridecommandlockouts
\usepackage{cite}
\usepackage{amsmath,amssymb,amsfonts}
\usepackage{algorithmic}
\usepackage{graphicx}
\usepackage{textcomp}
\usepackage{xcolor}
\usepackage{booktabs}
\usepackage{multirow}
\usepackage{bm}
\usepackage{enumitem}
\usepackage[justification=centering]{caption}
\def\BibTeX{{\rm B\kern-.05em{\sc i\kern-.025em b}\kern-.08em
    T\kern-.1667em\lower.7ex\hbox{E}\kern-.125emX}}
\begin{document}

\title{%Advancing Multi-Human Motion Generation through Interaction-Embedded Latent Space\\
%Unified Multi-Person Interaction Modeling for Motion Generation
%Representing Motion and Interaction Using Unified Representation of Two-Person Interaction for Multi-Person Motion Generation 
%Two-in-One: Unified Interaction Modeling for Multi-Person Motion Generation
Two-in-One: Unified Multi-Person Interactive Motion Generation by Latent Diffusion Transformer
% {\footnotesize \textsuperscript{*}Note: Sub-titles are not captured for https://ieeexplore.ieee.org  and
}

\author{
    \IEEEauthorblockN{
        Boyuan Li, 
        Xihua Wang, 
        Ruihua Song and
        Wenbing Huang
    }\\
    \IEEEauthorblockA{
        Gaoling School of Artificial Intelligence, Renmin University of China, Beijing 100872, China\\
        Email: 
            \{liboyuan, xihuaw, rsong\}@ruc.edu.cn, hwenbing@126.com
    }
}

\maketitle

\begin{abstract}
Multi-person interactive motion generation, a critical yet under-explored domain in computer character animation, poses significant challenges such as intricate modeling of inter-human interactions beyond individual motions and generating two motions with huge differences from one text condition.
Current research often employs separate module branches for individual motions, leading to a loss of interaction information and increased computational demands. 
To address these challenges, we propose a novel, unified approach that models multi-person motions and their interactions within a single latent space.
Our approach streamlines the process by treating interactive motions as an integrated data point, utilizing a Variational AutoEncoder (VAE) for compression into a unified latent space, and performing a diffusion process within this space, guided by the natural language conditions.
Experimental results demonstrate our method's superiority over existing approaches in generation quality, performing text condition in particular when motions have significant asymmetry, and accelerating the generation efficiency while preserving high quality. 

\end{abstract}

\begin{IEEEkeywords}
Latent Diffusion Model, Multi-modalities, Human interaction generation
\end{IEEEkeywords}

\section{Introduction}
% Human motion generation is a critical task in computer character animation, with controllable human motion generation finding broad applications in areas such as video games, animated film production, virtual/augmented reality, and robot control. The rapid development of generative models has injected new vitality into traditional character animation production processes; the addition of multimodal models allows for the use of cross-modal control signals, such as audio\cite{listen}, \cite{aistpp}, \cite{bailando}, \cite{audio2gestures}, \cite{beat}, action categories\cite{action2motion, ACTOR, ac, posegpt}, natural language\cite{humanml3d}, \cite{mdm}, \cite{text2action}, \cite{TEMOS}, \cite{tm2t}, even physical signals\cite{physdiff}, \cite{omnicontrol}. The use of the cross-modal information makes a more human-friendly motion generation experience possible\cite{motionclip}, \cite{tomato}, \cite{LDM}, \cite{emdm}, significantly reducing the threshold for animators to use and potentially greatly improving their production process, enhancing efficiency.

Human motion generation is crucial in computer character animation, with wide applications such as video games, animated films, and virtual reality. 
The emergence of deep generative models~\cite{gan, vae, flow, ddpm} has revolutionized conventional motion generation techniques. By conditioning on user-friendly input text, these generative methods can produce lifelike and realistic motion sequences, offering the potential to streamline animation workflows and improve efficiency~\cite{mdm, action2motion ,ACTOR, text2action}. Despite significant progress, most existing generative methods are limited to generating single-person motions, with multi-person scenarios remaining less explored, which is our focus.

\begin{figure*}[t]
\centering
\includegraphics[width=0.88\linewidth]{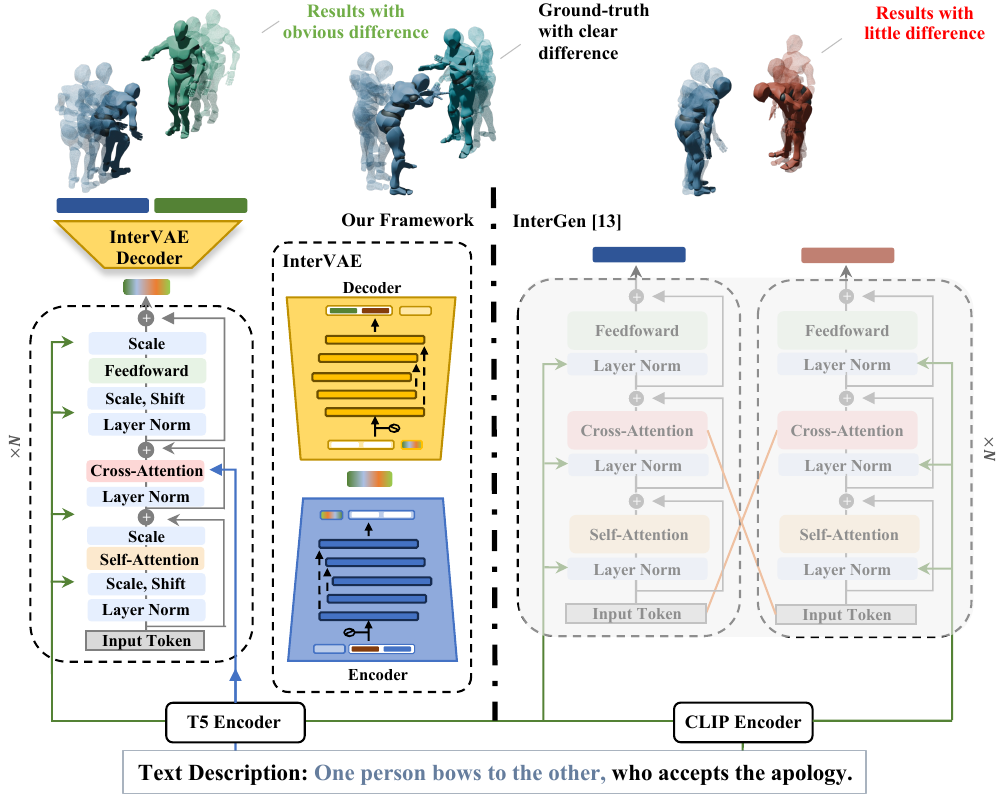}
\caption{Left: Our proposed framework, which uses an interaction Variational AutoEncoder (InterVAE) to encode two-person motions into a unified latent space and uses a conditional interaction latent diffusio(InterLDM) generate the latents.\\ Right: InterGen is the state-of-the-art work using two-branch framework  with cross-attention interactions to generate motions.}
\label{fig:model}
\end{figure*}

Compared to the single-person motion generation~\cite{text2motion, ACTOR, mdm, humanml3d}, 
multi-person interactive motion generation is more challenging because it demands both high-quality individual motion generation and accurate inter-motion communication and interaction. For example, as shown in Figure~\ref{fig:model}, given a text condition ``One person bows to the other, who accepts the apology.'', 
the model is expected to accurately assign motions to different persons and ensure that the catch neither precedes the toss nor occurs after the ball has landed.

% The Intergen\cite{intergen}mo-cap dataset, annotated with natural language, contains a considerable amount of two-person interactive motion data, bringing hope to the solution of the aforementioned problems. Based on conditional diffusion\cite{CDM}, attempts are made to incorporate the posterior distribution of the other person's motion into the denoising process using the cross-attention mechanism, and to control the content of the generated interactive motions with the corresponding natural language text description, improving the quality and diversity of two-person interactive motion generation. However, its scheme of performing the diffusion process on the raw data usually increases the redundant computational load, leading to low efficiency in training and inference; in terms of data representation, it still treats the dual-person motion data as two separate single-person motion sequences, which to some extent also leads to the loss of interaction information.

Previous works for multi-person motion generation usually leverage pre-trained single-person motion generation models, generating independent single-person motions and then integrating them via a motion-communication module~\cite{comMDM} or motion space control signals~\cite{intercontrol}.
Recent studies directly model multi-person motions from scratch, but they still rely on individual generative branches supplemented with a cross-attention module to model interactions~\cite{intergen}, or consider the interaction problem as a sequential generation problem~\cite{freemotion}.
These approaches separate multi-person motion modeling into individual motion modeling, potentially leading to a loss of interaction information, such as relative spatial relationships. As shown in Figure~\ref{fig:model}, the state-of-the-art model~\cite{intergen} generates the right motion of a person bowing, but it fails to generate another person facing the first person and accepting the apology. In addition, taking the description as a common condition of two individual branches may be another cause of such failures when the motions of two persons has significant asymmetry.% and expressing, such as assigning both tossing and catching to one person while the other remains idle. These practices have yet to adequately address the persistent challenges in multi-person motion generation, often leading to ``low interaction density'', as illustrated in Figure~\ref{fig:model}.

In this paper, we propose a unified framework to model the multi-person interactive motions by latent diffusion transformer.%and inter-human interactions within a single latent space, directly using one generative network branch rather than multiple independent branches to generate multi-person motions. 
%Our key insight is that, by taking dual-person motions for example, unified motions retains more comprehensive information exceeding the sum of individual motions and their interactions, which implies the inequality: $\text{Motion}_{\text{dual-person}} > \text{Motion}_1 + \text{Motion}_2 + \text{Interaction}$.
First, to maintain full information of individual motions and inter-human interaction, we propose treating two-person motions as a single data point and then employing a Variational AutoEncoder (VAE)\cite{vae} to compress each data point into one latent space for better representation. 
Then we propose learning a single diffusion generative network based on this unified latent space while using input text as condition to guide the entire two-person motions. As shown in Figure~\ref{fig:model}, our proposed ``two-in-one'' motion modeling can correctly generate one person bows to the other, who accepts the apology. In addition, our method is theoretically more efficient compared to previous two-branch design because the unified latent space can compress the motion sequence into less than a tenth of the original length and a unified diffusion model also does not double the sequence length during generation.
Experimental results indicate that our proposed approach outperforms all baselines in R-Precision, FID, and Diversity.
%Aside from generation quality, multi-person motion generation significantly increases the computational demands compared to single-person motion generation. 
%Current dual-person motion generation methods model motion sequences in the raw motion data space~\cite{intergen}, where a motion sequence can reach a length of 256. Due to the multi-branch model design, generating motions for two people requires handling a sequence length of 512. 
%Our method first compresses the motion sequence to less than a tenth of the original raw sequence length, from a sparse data space to a compact latent space. 
%And with a two-in-one design, dual-person motion generation does not necessitate doubling the sequence length. 
%Instead, a singular network generates samples within this compact, unified latent space, thereby substantially improving generation efficiency while preserving quality.

Our contributions can be summarized as follows:
\begin{itemize}[leftmargin=*]
\item To our best knowledge, we are the first to utilize VAE to encode multi-person interactive motions into a unified latent space with less information loss and better expressiveness.% for multi-person interactive motions, bringing significant benefits to the subsequent generation task.
% \item We employ a latent diffusion model to characterize the latent space representations of interaction, concurrently integrating dual textual features to precisely delineate the interplay between the distributions of textual representations and those of interactive motions.
\item We propose employing one latent diffusion transformer to fit the distribution of multi-person interactive motions and better take advantage of multi-person motion descriptions.%the text condition guide approach by directing the unified interactive motions, rather than separate individual motions.
% \item We have identified our approach that not only achieves the most efficient and accurate method for generating interactive motions but also consistently yields optimal results across relevant metrics. 
\item Extensive experimental results indicate that our method consistently surpasses all baselines in terms of generation quality and diversity, while accelerating the generation speed by more than 4x while maintaining elevated quality.
% \item We are the first to propose and use a method for evaluating the quality of interactive motions, which can conveniently and accurately assess the strength of the relationship between two people's motions, bringing great convenience to subsequent work in this field.
\end{itemize}

\begin{table*}[htb]
\caption{
\fontsize{9pt}{11pt} \selectfont \MakeUppercase {
Quantitative evaluation results on the InterHuman test set
}
\\ $\pm$ indicates the 95\% confidence interval. \textbf{Bold} indicates best result} \label{tab:main}
\begin{center}
\begin{tabular}{c|c|c|c|c|c|c|c}
\toprule
\multirow{2}{*}{\textbf{Methods}}&\multicolumn{3}{|c|}{R Precision \(\uparrow\)} & \multirow{2}{*}{FID \(\downarrow\) } & \multirow{2}{*}{MM Dist  \(\downarrow\)} & \multirow{2}{*}{Diversity} & \multirow{2}{*}{MModality} \\
 % \cline{2-4} 
 & Top1& Top2& Top3 & & & &\\
 % \cline{2-8} 
\midrule
Real & $0.452^{\pm 0.008}$ & $0.610^{\pm 0.007}$& $0.701^{\pm 0.008}$ &$0.273^{\pm 0.007}$ &  $3.755^{\pm 0.008} $& $7.948^{\pm 0.064}$ & - \\
\midrule
TEMOS\cite{TEMOS} & $0.224^{\pm 0.010}$ & $0.316^{\pm 0.013}$& $0.450^{\pm 0.018} $&$17.375^{\pm 0.043}$ &  $6.342^{\pm 0.015}$ & $6.939^{\pm 0.071}$ & $0.535^{\pm 0.014}$ \\
T2M\cite{text2motion} & $0.238^{\pm 0.012}$ & $0.325^{\pm 0.010}$& $0.464^{\pm 0.014}$ &$13.769^{\pm 0.072} $&  $5.731^{\pm 0.013}$ & $7.046^{\pm 0.022} $& $1.387^{\pm 0.076}$ \\
ComMDM\cite{comMDM} & $0.223^{\pm 0.009}$ & $0.334^{\pm 0.008}$& $0.466^{\pm 0.010}$ &$7.069^{\pm 0.054}$ &  $6.212^{\pm 0.021}$ & $7.244^{\pm 0.038}$ & $1.822^{\pm 0.052}$ \\
FreeMotion\cite{freemotion} & $0.326^{\pm 0.003}$ & $0.462^{\pm 0.006}$ & $0.544^{\pm 0.006}$ & $6.740^{\pm 0.130}$ &$ 3.848^{\pm 0.002}$ & $7.828^{\pm 0.130}$ & $1.226^{\pm 0.046}$ \\
InterGen\cite{intergen} &$0.371^{\pm 0.010}$& $0.515^{\pm 0.012}$& $0.624^{\pm 0.010}$&$5.918^{\pm 0.079}$&  $5.108^{\pm 0.014} $&$7.387^{\pm 0.029}$ &$\textbf{2.141}^{\pm 0.063} $ \\
\midrule
InterLDM(Ours) & $\textbf{0.427}^{\pm 0.004}$ & $\textbf{0.559}^{\pm 0.05}$& $\textbf{0.638}^{\pm 0.004}$ & $\textbf{5.619}^{\pm 0.091}$& $ \textbf{1.862}^{\pm 0.007}$& $\textbf{7.888}^{\pm 0.041}$ & $1.032^{\pm 0.089} $\\
\bottomrule
% \multicolumn{4}{l}{$^{\mathrm{a}}$Sample of a Table footnote.}
\end{tabular}
\label{tab1}
\end{center}
\end{table*}

\section{Method}

% 总起：
% 第一段，task formulation，不用展开x_i^t = [j_p, j_v, j_r, b_{fc}], 用文字替代，简述。说一下follow的哪。
% 第二段，commen的解法是在上述数据空间中用双支网络建模双人动作，as shown in Fig.1。这样做低效且interaction效果不好。如图1，我们用VAE将两个人动作塞进latent空间既降低维度又保留interaction信息，然后再用一支网络建模interaction，高效且强interaction。我们将在A中介绍VAE，B中diffusion，C中它们的训练过程和最终的推理过程。

% A。VAE：
% 补充1：强调这个部分是干嘛的，将x1 x2 => z。
% 补充2：是用序列更短的learnabel的token压缩原序列到 latent，这点。

% B。Generator：
% 补充1：强调这个部分是干嘛的，given第一步压缩好的z，建模z的分布。

% C。training and inference。
% A和B只负责讲到L的定义为止，整个优化过程的规划在这讲。A中有L1，B中有L2，整个优化过程是先L1再L2。
% inference。infer的过程 z中采样，再VAE解码，巴拉巴拉。以及CFG guidance。可以强调一嘴单支网络、单次condition输入。
%In this section, we first formulate the problem, overview our framework and describe main components in details respectively.

Given a text description \( c = \{w^{1:N}\} \) as input, our goal is to generate a sequence with \( L \) frames of pairwise interactive motions \( x = \{x^{1:L}_1, x^{1:L}_2\} \) that matches the description. We employ a non-canonical representation\cite{intergen} for individual motions 
% \(x_i^t = [j_p, j_v, j_r, b_{fc}]\)
, which allows us to directly access the joint positions of all individuals in the world frame, facilitating the capture of their relative relation. It also provides abundant data, such as joint rotations and foot-ground contact, which aids the model in learning the motion patterns of individuals.

%However, directly modeling from this pairwise motion data would result in exponentially increasing computational costs. On the one hand, adapting the training paradigm from single-person motion data to two-person motion sequences is challenging and leads to a loss of interactive information if split into single-person processing. On the other hand, the low information density of human motion sequences also has negative effects on training and inference. 
We propose a unified framework to solve the problem, as shown in Figure~\ref{fig:model}. To efficiently generate paired motion sequences while retaining complete interaction information, we first design an Interaction Variational AutoEncoder (InterVAE) in Section~\ref{S2-1} to map the entire multi-person interactive motions into a low-dimensional latent space. Then, given the text description, we propose a conditional Interaction Latent Diffusion Model (InterLDM) in Section~\ref{S2-2} to capture the relationship between the text condition and the embedding of interactive motions in our InterVAE latent space. In addition, we provide the training and inference details in Section~\ref{S2-3}.

%In this Section, we will present the latent space construction proceeding using Interaction Variational AutoEncoder(InterVAE) in \ref{S2-1}. In \ref{S2-2}, we present the specific implementation of the conditional Interaction Latent Diffusion Model(InterLDM). Lastly, we provide the training and inference details in \ref{S2-3}.

\subsection{Interaction Variational Autoencoder} \label{S2-1}

We propose an interaction Variational AutoEncoder called InterVAE, which consists of a transformer encoder $\mathcal{E}$ and a transformer decoder $\mathcal{D}$, to encode multi-person interactive motions into one latent space. The interaction encoder \(\mathcal{E}\) takes normalized interactive motions \(x\) concatenated with a learnable latent token as inputs. The embedded token output by the encoder is used to re-parameterize the low-dimensional latent variable \( z \in \mathbb{R}^{n\times d} \) for \( x \) embedded in the latent space \(\mathcal{Z}\). Then the interaction decoder \(\mathcal{D}\) takes the latent variable as the head of the zero interactive motion sequences, performing the reconstruction process and generating the interaction sequence \( \hat{x}\).
In our experiments, the latent variable size is about \textbf{1/10} of the raw data when achieving the best reconstruction results.

In the objective function, we incorporate not only commonly used losses for VAE training, such as Mean Squared Error(MSE) and Kullback-Leibler(KL) divergence loss, but also several optimization objectives tailored to the geometric features of the human body, including foot contact, bone length, and joint velocity loss, which ensure that the reconstruction process accurately reflects the dynamic structure of the human skeleton.
\begin{equation}
    L_{\text{VAE}} = L_{mse} + L_{kl} + L_{vel} + L_{bone} + L_{fc}.
\end{equation}

\subsection{Conditional Interaction Latent Diffusion Model} \label{S2-2}

%Learning the prior distribution of noise, diffusion probabilistic models can gradually approximate the true data distribution from noise sampled from a standard Gaussian distribution. Thus, 
Based on on the unified interaction latent space constructed by the InterVAE, we propose a conditional interaction latent diffusion model called InterLDM to learn the prior distribution of noise and approximate the real data distribution from sampled noise. As multi-person interaction sequence is a kind of sequential data, we use Diffusion Transformer(DiT)\cite{dit} as the denoiser \(\epsilon_{\theta}\), which proves to be effective in generating long sequences of images.
Specifically, we first get the embedded interaction latent variable \(z\) from the frozen InterVAE Encoder \(z_0 = \mathcal{E}(x)\), and then we perform the forward diffusion process\cite{ddpm} on the latent variable as:

% However, training diffusion models directly on the original data sequence of two-person motions incur substantial computational costs. On the one hand, adapting the training paradigm from single-person motion data to two-person motion sequences is challenging and may lead to a loss of interactive information if split into single-person processing. On the other hand, the information density of human motion sequences is relatively low. To address these issues, we propose performing the diffusion process in a high-information-density latent space, as constituted by the aforementioned VAE. This approach not only significantly reduces the computational requirements but also preserves interactive information, thereby enhancing the quality of the generated motions.

\begin{equation} \label{eq:forward_diffusion}
    q(\bm{z}_t|\bm{z}_{t-1}) = \mathcal{N} (\sqrt{\alpha}_t \bm{z}_{t-1}, (1-\alpha_t)I), 
\end{equation}
where the constant \(\alpha\) is a hyper-parameter for sampling, \(\bm{z}_t\) is the latent variable representation of the sequence \(x\) encoded by the encoder and added with noise \(\epsilon_t\) at the \(t\)-th step. 

At t-step in the reverse diffusion process, our denoiser predicts the noise distribution \(\epsilon_{\theta}(\bm{z}_t, t, c)\) conditioned by the text description embedding \(c\) using the AdaLN-Zero~\cite{dit} mechanism and then removes this noise from the current latent variable to get the previous one, denoted as follows:

\begin{equation}
    L_{\text{LDM}} = \mathbb{E}_{\epsilon,t,c\sim \mathcal{N}(0,1)}[||\epsilon - \epsilon_\theta(\bm{z}_t, t, c)||_2^2].
\end{equation}
Once the reverse process is complete, we can get the predicted noise-free variable \(\hat{z_0}\) for subsequent decoding usage.

Our denoiser uses the classifier-free guidance method\cite{cfg} to learn both conditional and unconditional distributions by randomly masking 10\% of the description content. This approach allows us to balance diversity and fidelity by adjusting the classifier-free guidance scale \(s\) when inference:
\begin{equation}
    \epsilon_\theta(\bm{z}_t, t, c) = s \epsilon_\theta(\bm{z}_t, t, c) + (1-s)\epsilon_\theta(\bm{z}_t, t, \emptyset).
\end{equation}

\subsection{Training and Inference} \label{S2-3}

We first train the InterVAE on the reconstruction task to obtain a reasonable low-dimensional latent space for high-density interactive motion representation. The interaction encoder and decoder all consist of 9 layers and 4 heads with skip connections. Furthermore, we employ both a frozen \textit{CLIP-ViT-L-14}\cite{clip} and a frozen \textit{T5-small}\cite{T5} text encoder for encoding the conditional description to word-level and sentences-level embeddings like \cite{SD3}. %, injecting which by AdaLN-Zero\cite{dit} and cross-attention separately. 
With these conditional inputs, we train the denoiser \(\epsilon_\theta\) of 12 DiT blocks , starting from the compressed latent variable \(\bm{z}_0\) and sampling a \(\bm{z}_t\) as (\ref{eq:forward_diffusion}) 
 does for denoiser learning to predict the \(\bm{z}_{t-1}\) on random t-step.

During the inference phase, our InterLDM performs a single diffusion process on a latent variable randomly sampled from Gaussian noise. %rather than using the previous method which involved reverse diffusion on two branches with heavy raw-data-size variable. 
By employing DPMSolver++\cite{dpmsolver} as the noise scheduler, after 25 denoising steps, we obtain a noise-free latent variable conditioned by the given description, which is then fed into InterVAE's decoder for reconstructing human interactions, generating full interactive human motion data.

\section{Experiment}
\subsection{Experiment Setup}
\textbf{Dataset} Our model is trained on a public large-scale human motion capture dataset, \textbf{InterHuman}\cite{intergen}, which contains 6,022 instances of two-person interactive motions with strong interaction intensity, with a total duration of 6.56 hours. Each interactive motion instance is accompanied by approximately three human-annotated natural language descriptions. %To our best knowledge, this is currently the largest dataset of two-person interactive motions annotated with natural language.

\textbf{Evaluation} To comprehensively evaluate the performance of the latent diffusion model, we adopt the following metrics widely used in previous works~\cite{intergen}: 1) \textbf{Recognition Precision} evaluates the similarity between text-motion data pairs; 2) \textbf{FID} measures the difference between the distribution of raw motion embeddings and the generated motion embeddings; 3) \textbf{MM Dist} measures the distance between the embeddings of generated motion sequences and text embeddings; 4) \textbf{Diversity} assesses the distance between the generated motion sequences; 5) \textbf{MModality} evaluates the distance between motion sequence embeddings generated under the same text conditions. 
% Since these text-motion embedding pairs are close in geometric space, we can use these embeddings to evaluate from the following perspectives:

\begin{figure}[t]
\centering

\includegraphics[width=\linewidth]{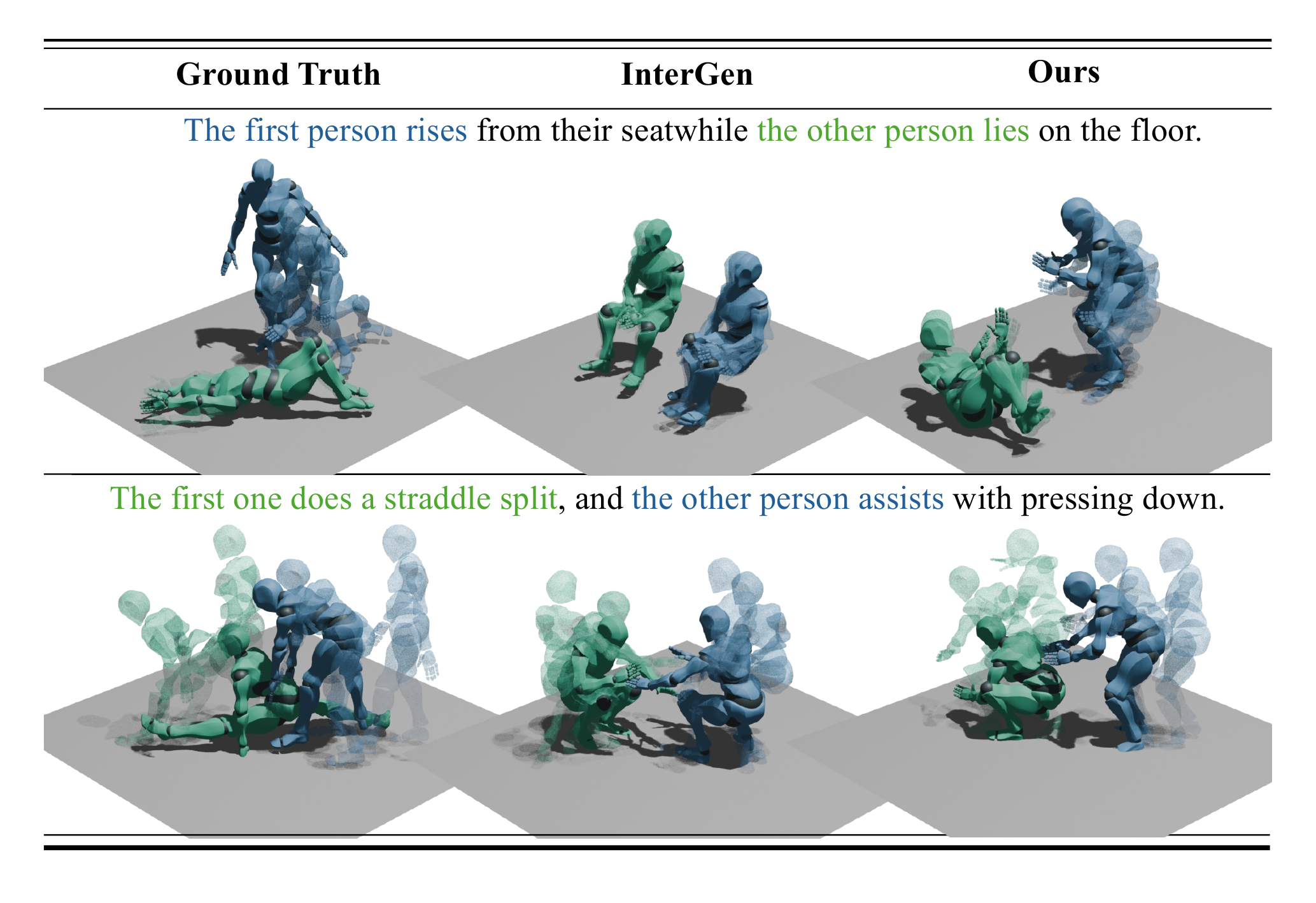}
\caption{Visualization of generated interactive motions from InterGen\cite{intergen} and Ours.}
\label{fig:cases}
\label{cases}
\vspace{-0.5cm}
\end{figure}

\subsection{Compare with Baselines}

We compare our proposed method with baselines including the single-person methods VAE-based TEMOS\cite{TEMOS} and T2M\cite{text2motion}, diffusion-based seperated-branch methods ComMDM\cite{comMDM}, FreeMotion\cite{freemotion}, and InterGen\cite{intergen} and present results in Table~\ref{tab:main}. %The VAE-based network are adjusted on input and output dimensions to fit interaction representation, most results are borrowed from \cite{intergen} except the inference time for the absence of the running time in their own paper.
%testing 20 times at a 95\% confidence level. 
As the results shows, our InterLDM model performs the best in terms of R Precision, FID, MM Dist, and Diversity. This indicates that our model has the capability to generate high-quality and highly-relevant multi-person interactions. The improvements in R Precision and MM Distance are the most significant, indicating our method can significantly improve the ability to follow text conditions or instructions. In addition, we observe that our method's Diversity is closest to ground-truth while its MModality, measuring the diversity under the same text condition, is lower than InterGen and other three baselines. This indicates that for different text instructions, our method can generate the most diverse motions, while given a text condition, our method generate more less diverse motions. However, if the generated motions are not relevant to the text condition, the higher MModality means the worse instruction-following ability.

We conduct a case study and show some cases in Figure \ref{fig:cases}. First, we observe that our InterLDM method can generate more accurate motions, which are more like the ground-truth, under the conditional textual descriptions. Second, the two persons in InterGen generated results tend to do the similar actions, e.g., both rises in the first case and both in the third case, which confirms that it is difficult for two separate individual branches with interaction via cross-attention mechanism to generate two persons' motions with huge difference under the same text condition. In contrast, our InterLDM model effectively preserves the asymmetrical interactive information between the two individuals and generates results correctly following the text instruction.

\begin{figure}[t]
\centering \vspace{-0.5cm}
\includegraphics[width=0.85\linewidth]{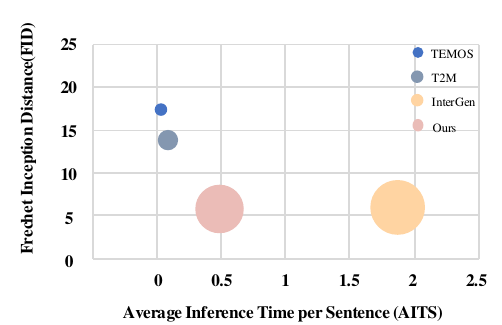} 
\caption{Comparison of inference time, FID and quantity of parameter. All tests are performed on the same Tesla A100.}
\label{fig:bubble}
\label{inference-speed}
\end{figure}

We also compare some methods from three perspectives: the number of parameters (by dot size), FID and average inference time per sentence. The results shown in Figure~\ref{fig:bubble} indicate that although some small models like TEMOS and T2M are faster, their generation quality is much worse than InterGen and InterLDM. With similar size, our InterLDM model is about 4x faster than InterGen while achieving better quality.

%\subsection{Efficiency}

%Another notable improvement offered by our InterLDM is its ability to decrease training and inference computational costs commonly associated with diffusion models. 
%As shown in Figure~\ref{fig:bubble}, we compare the average inference times per sentence and FID of four methods. TEMOS and T2M are fasour method significantly reduces the costs by utilizing VAE to increase the information density of interaction data. Using DDIM\cite{ddim} as a sampler for 50-step inference on a single V100 GPU. 
%The inference times of other models range from $0.029$ to $1.868$ seconds, while our model completes the process in $0.487$ seconds and achieves superior quality results.

\begin{table}[t]
\caption{Reconstruction and generation performance with different token length f.} \label{tab:ablation}
\begin{center}
\begin{tabular}{c|c|c|c}
\toprule
\textbf{\( z \in \mathbb{R}^{f\times 256}\)} & $\text{FID}_{\text{Recon}}$ & $\text{FID}_{\text{Gen}}$ & Average Inference Time(s)\\ %& R Precision Top3\(\uparrow\) \\

\midrule
f=18 & 1.970 & 7.323 &  \textbf{0.424}\\
f=24 & \textbf{1.330}& \textbf{5.619} & 0.487\\
f=30 & 1.892 & 6.221 & 0.557\\
f=36 & 1.919 & 8.273 & 0.584\\
% \midrule
% \midrule
% IMLD(\( z \in \mathbb{R}^{18\times 256}\)) & - & -\\
% IMLD(\( z \in \mathbb{R}^{24\times 256}\)) & - & -\\
% IMLD(\( z \in \mathbb{R}^{30\times 256}\)) & - & -\\
% IMLD(\( z \in \mathbb{R}^{36\times 256}\)) & - & -\\
\bottomrule
% \multicolumn{4}{l}{$^{\mathrm{a}}$Sample of a Table footnote.}
\end{tabular}
\label{tab2}
\end{center}
\vspace{-0.5cm}
\end{table}

\subsection{Ablation Study}
The unified latent space plays an important role in our framework and thus we explore the reconstruction effects of \( z \in \mathcal{R}^{f \times 256} \) on both the InterVAE and the diffusion denoiser. Here, a smaller \( f \) implies a higher compression rate and a more compact latent space. Table~\ref{tab:ablation} shows the model's reconstruction and generation performance for different values of \( f \). We find that higher compression rates lead to faster inference but can degrade reconstruction quality, highlighting a trade-off between inference speed and reconstruction quality. Furthermore, lower compression rates can negatively impact the model's generation performance, indicating that the sparsity of interaction information in the raw data space adversely affects generation results. Therefore when $f=24$, InterLDM achieves high-quality synthesis results with impressive speed.

% InterLDM consists of a VAE model responsible for encoding and decoding interactive motions and a diffusion model responsible for aligning textual information with interactive representations.
%We first tested the reconstruction effect of the VAE under different parameters, and then based on the optimal VAE, we conducted generation and speed tests on the diffusion model.

\section{Conclusion}
To generate multi-person interaction motions, we propose an interaction Variational AutoEncoder to effectively preserves the full information to two-person interaction motions, and a conditional latent diffusion model to efficiently generate two-person motions together and ensure consistency with textual description. Extensive experiments demonstrate that our model not only performs the best in generation quality and instruction-following ability, but also achieves a 4x faster inference speed than strong baselines. In the future, we plan to explore more unified framework that can model interactive motion generation for any number of persons.
% The proposed interaction matching rate metric provides a quantitative assessment of the interactivity quality, which is a significant contribution to the field. 

% \newpage
\clearpage

\bibliographystyle{IEEEbib}
\bibliography{ref}

\end{document}